\documentclass[11pt]{article}

\usepackage[preprint]{templates/acl2026/latex/acl}

\usepackage{times}
\usepackage{latexsym}

\usepackage[T1]{fontenc}

\usepackage[utf8]{inputenc}

\usepackage{microtype}

\usepackage{inconsolata}

\usepackage{graphicx}
\usepackage{float}

\usepackage{hyperref}
\usepackage{url}

\usepackage{graphicx}
\usepackage{caption}
\usepackage{subcaption}
\graphicspath{{manuscript/figures/}}

\usepackage{algorithm}
\usepackage{algorithmic}

\usepackage{mathtools}  
\usepackage{amssymb}
\usepackage{amsthm}

\usepackage[many]{tcolorbox}
\usepackage{listings}
\usepackage{enumerate}
\usepackage{enumitem}

\usepackage{booktabs}
\usepackage{multirow}
\usepackage{makecell}

\usepackage{listings}
\usepackage{xcolor}
\usepackage[dvipsnames]{xcolor}

\usepackage{xcolor}
\usepackage{soul}
\usepackage[textwidth=3cm, textsize=scriptsize]{todonotes}

\newcommand{\zhiyu}[1]{\textcolor{black}{#1}}

\newcommand{\ourmethod}{Draft-and-Prune}
\newcommand{\ourmethodabbr}{D\&P}

\newcommand{\VerbGen}{draft}

\newcommand{\Acc}{\ensuremath{\mathbf{Acc}}}

\newcommand{\AccAF}
{\ensuremath{\mathbf{Acc}_{\mathrm{AF}}}}

\newcommand{\ExecRate}{\ensuremath{\mathbf{ExecRate}}}

\newcommand{\AccAtExec}{\ensuremath{\mathbf{Acc@Exec}}}

\newcommand{\AccAFOnLSATOurGPTFour}{78.43\%}

\newcommand{\AccAFOnLSATOurGPTFourO}{78.00\%}
\newcommand{\AccOnLSATOurGPTFourO}{82.83\%}

\title{\ourmethod{}: Improving the Reliability of Auto-formalization for Logical Reasoning}

\author{
\textbf{Zhiyu Ni}$^{1}$\thanks{Equal contribution.},
\textbf{Zheng Liang}$^{1}$\footnotemark[1],
\textbf{Liangcheng Song}$^{2}$\footnotemark[1],
\textbf{Chenrui Cao}$^{2}$,
\textbf{Xian Zhang}$^{2}$, \\
\textbf{Alberto Sangiovanni-Vincentelli}$^{1}$,
\textbf{Pierluigi Nuzzo}$^{1}$ \\
$^1$University of California, Berkeley \quad
$^2$Microsoft \\
{\tt\small \{zhiyuni, zhliang, alberto, nuzzo\}@berkeley.edu,} \\
{\tt\small \{v-liasong, v-chenruicao, zhxian\}@microsoft.com}
}

\begin{document}

\maketitle

\begin{abstract}

Auto-formalization (AF) translates natural-language reasoning problems into solver-executable programs, enabling symbolic solvers to perform sound logical deduction.
In practice, however, AF pipelines are currently brittle: programs may fail to execute, or execute but encode incorrect semantics.
\zhiyu{While prior work largely mitigates syntactic failures via repairs based on solver feedback, reducing semantics failures remains a major bottleneck. 
}
We propose \ourmethod{} (\ourmethodabbr{}), an inference-time framework that improves AF-based logical reasoning via \emph{diversity} and \emph{verification}. 
\ourmethodabbr{} first \emph{drafts} multiple natural-language plans and conditions program generation on them.
It further \emph{prunes} executable but contradictory or ambiguous formalizations, and aggregates predictions from surviving paths via majority voting.
Across four representative benchmarks (AR-LSAT, ProofWriter, PrOntoQA, LogicalDeduction), \ourmethodabbr{} substantially strengthens AF-based reasoning without extra supervision.
\zhiyu{On AR-LSAT, in the AF-only setting, \ourmethodabbr{} achieves \AccAFOnLSATOurGPTFour{} accuracy with GPT-4 and \AccAFOnLSATOurGPTFourO{} accuracy with GPT-4o, significantly outperforming the strongest AF baselines MAD-LOGIC and CLOVER. 
}
\ourmethodabbr{} then attains near-ceiling performance on the other benchmarks, including 100\% on PrOntoQA and LogicalDeduction.\footnote{Code repository: \url{https://github.com/zyni2001/draft-and-prune}}

\end{abstract}

\section{Introduction}

\begin{figure*}[t]
     \centering
     \includegraphics[width=\linewidth]{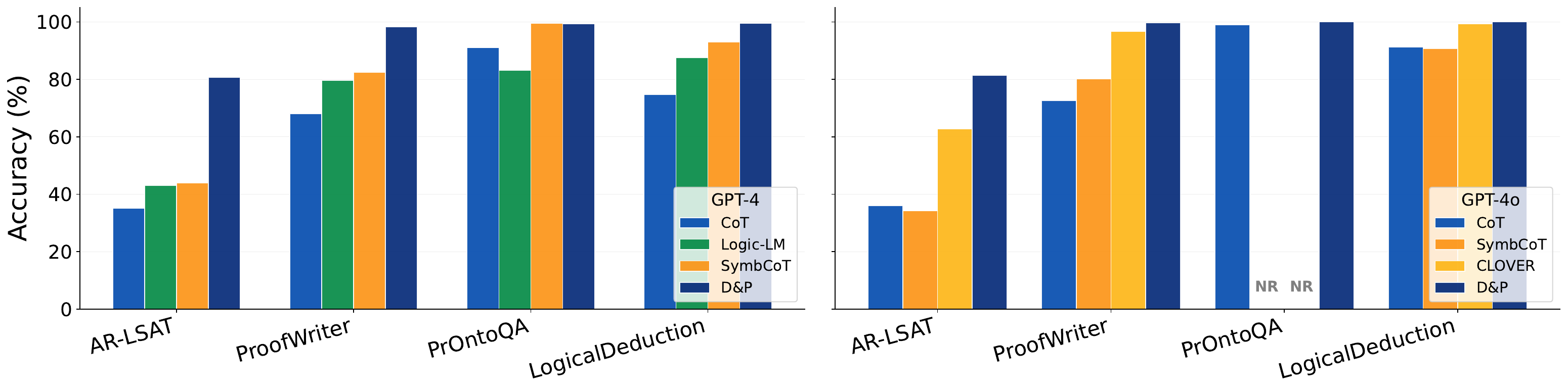}
     \caption{End-to-end accuracy (\Acc{}) on four logical reasoning benchmarks. Left: GPT-4 results comparing CoT, Logic-LM, SymbCoT, and \ourmethodabbr{}. Right: GPT-4o results comparing CoT, SymbCoT, CLOVER, and \ourmethodabbr{}. ``NR'' denotes not reported results. }
\label{fig:results}
\end{figure*}


Large language models (LLMs) achieve strong performance on many language tasks~\citep{achiam2023gpt,hurst2024gpt}, yet they remain unreliable on deductive logical reasoning~\citep{ar-lsat, logicaldeduction}.
Symbolic solvers provide sound deductive inference, but they require inputs written in precise formal languages (e.g., logic programs) rather than natural language.
Neuro-symbolic methods~\citep{zhang-etal-2023-improved} aim to combine the flexibility of LLMs with the rigor of symbolic reasoning.
Among them, \emph{auto-formalization} (AF) translates a natural-language problem into an executable formalization that a solver can run, thereby offloading deduction to a sound backend~\citep{wu2022autoformalization, logic-lm}.

In practice, however, existing AF pipelines are brittle and often fail in two ways: (i) due to \emph{syntactic failures}, when the generated formalization cannot be parsed or executed, and (ii) due to \emph{semantic unfaithfulness}, where the program can be executed but does not capture the original intent.
Prior systems~\citep{logic-lm} largely mitigate syntactic failures by using the solver feedback (e.g., error messages) to repair the syntax, but semantic failures remain common. 
\zhiyu{We indeed observe that current AF frameworks tend to under-utilize the reasoning capabilities of the symbolic solver. By generating only a narrow set of candidate formalizations (often one plus a few repaired versions), they end up insufficiently exploring the space of possible formalizations. 
Increasing the number of sampled candidates can help raise the chance of discovering a correct formalization. However, na\"ive sampling, if not appropriately biased, may not be sufficient to significantly improve semantic faithfulness. 
} 

We address these limitations by proposing \ourmethod{} (\ourmethodabbr{}), an inference-time framework that:
(i) \emph{drafts} multiple high-level plans and conditions program generation on them;
(ii) \emph{prunes} executable candidates that are not well-defined (e.g., contradictory or ambiguous); and
(iii) aggregates the answers from the remaining candidates.



\ourmethodabbr{} improves the performance by over 30\% on AR-LSAT~\citep{ar-lsat}, a benchmark where existing auto-formalization baselines perform poorly. On other datasets, including PrOntoQA~\citep{prontoqa} and LogicalDeduction~\citep{logicaldeduction}, \ourmethodabbr{} achieves near-ceiling accuracy.  
Overall, these results suggest that auto-formalization-based reasoning can be a reliable approach when combined with inference-time diversity and pruning.

\noindent\textbf{Contributions.}
(i) We analyze the brittleness of existing AF pipelines and characterize key failure modes.
(ii) We propose \ourmethod{} (\ourmethodabbr{}), an inference-time framework that improves reasoning accuracy via draft-induced diversity and verification-based pruning.
(iii) We evaluate \ourmethodabbr{} on multiple deductive reasoning benchmarks, by reporting the end-to-end accuracy and other diagnostic metrics, and providing in-depth ablations and error analysis.

\section{Background}
\label{sec:background}

\subsection{Deductive logical reasoning}
\label{sec:deductive-reasoning}

Deductive logical reasoning derives conclusions that follow necessarily from a given set of premises.
Typical tasks include checking argument validity, proving theorems, and solving constraint problems.
Such tasks demand both syntactic precision and semantic consistency, and differ from inductive predictions, where machine learning algorithms dominate and statistical generalization is the focus.

Formally, a set of premises \(P\) is given together with one or more candidate hypotheses \(Q_i\).
The goal is to determine the logical relationship between \(P\) and each \(Q_i\), such as entailment (\(P \models Q_i\)), contradiction (\(P \models \lnot Q_i\)), or consistency (i.e., \(P \land Q_i\) is satisfiable).

\subsection{Reasoning with LLM and prompting}

Benchmarking studies in natural-language reasoning suggest that direct generation by LLMs can produce outputs that are fluent yet logically invalid~\cite{logicaldeduction, tafjord2021proofwriter}.
Prompting-based methods provide procedural scaffolding to improve accuracy and reliability.
Chain-of-Thought (CoT) prompting encourages intermediate reasoning steps~\citep{wei2022chain}, while self-ensembling (e.g., self-consistency) samples multiple reasoning traces and aggregates answers to improve accuracy and robustness~\citep{wang2022self}.
More recent methods explicitly structure the reasoning process with decomposition and planning, such as least-to-most prompting \citep{zhou2023least} and plan-and-solve prompting \citep{wang-etal-2023-plan}.
These ideas motivate our use of \emph{drafted plans} as lightweight scaffolding to diversify candidates.

\subsection{Reasoning based on symbolic solvers}



Despite increasingly sophisticated prompting strategies, LLMs can be unreliable on tasks that require strict logical validity, since their training objective provides no guarantee of sound deduction.
This has motivated neuro-symbolic methods that integrate neural generation with symbolic solvers~\citep{zhang-etal-2023-improved}.
A primary paradigm in this space is \emph{auto-formalization} (AF), which translates natural-language problem statements into solver-executable formalizations (e.g., logic programs illustrated in Figure~\ref{fig:case-z3py-correct}), that can be executed by symbolic solvers~\citep{logic-lm}.


In AF pipelines, failures can arise either from non-executable outputs (syntactic issues) or from executable but semantically incorrect encodings.
Prior work typically reports diagnostic metrics that separate (i) whether a generated solver program is executable (\emph{execution rate}) from (ii) whether executable programs yield the correct answer (\emph{accuracy among executable formalizations}) to disentangle syntactic/tooling failures from semantic unfaithfulness.
We follow this convention in our analysis; formal definitions are given in Section~\ref{sec:metrics}.

\begin{figure*}[t]
    \centering
    \includegraphics[width=0.8\linewidth]{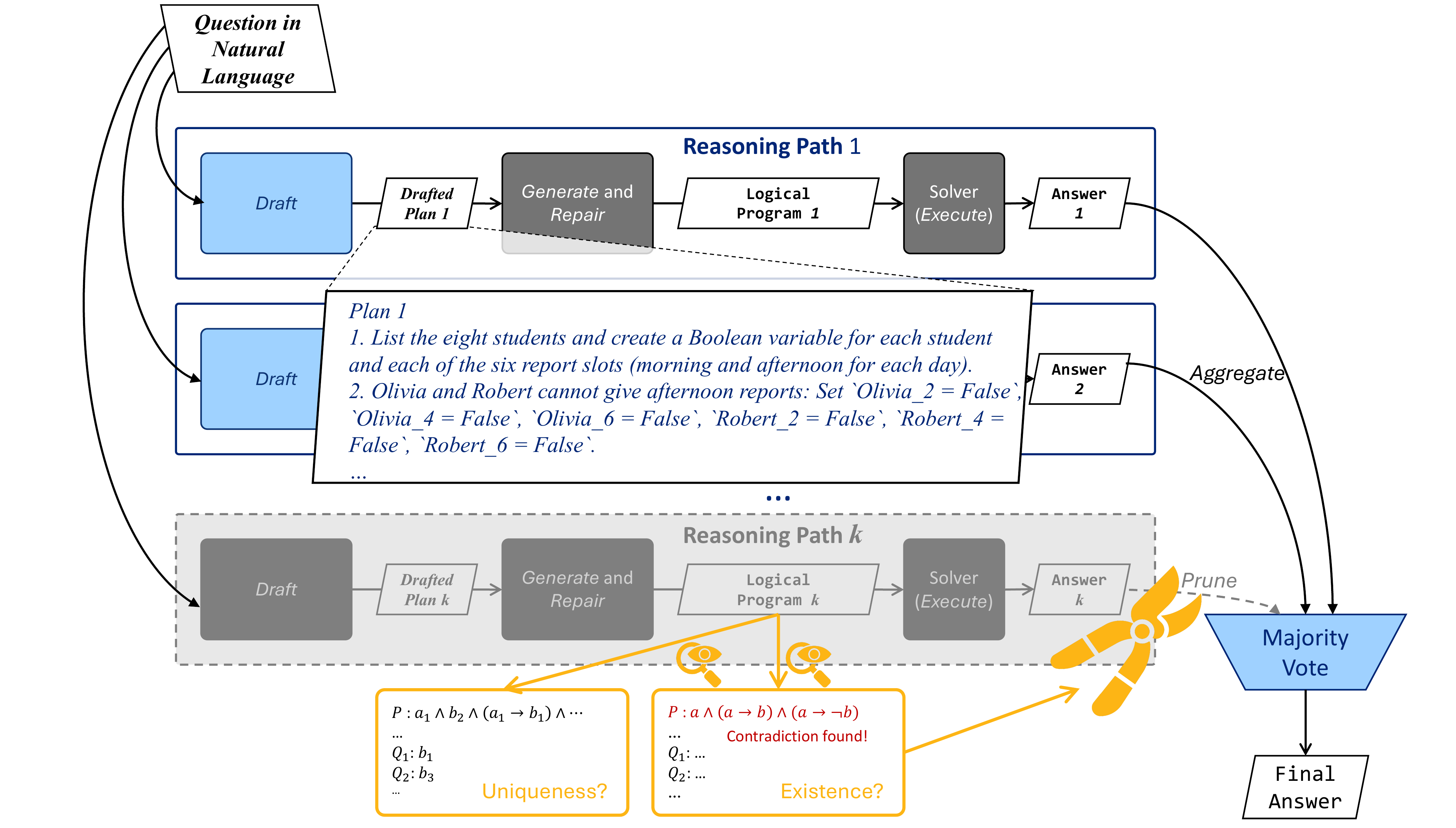} 
    \caption{The algorithmic pipeline of \ourmethod{} }
    \label{fig:algorithm-pipeline}
\end{figure*}

\section{\ourmethod{}}
\label{sec:method}

\begin{figure}[t]
    \centering
    \input{manuscript/case-lsat-z3py-correct.tex}
    \caption{A correct Z3Py program corresponds to an input question in AR-LSAT}
    \label{fig:case-z3py-correct}
\end{figure}

\begin{figure}[t]
    \input{manuscript/case-lsat-z3py-contradiction.tex}
    \input{manuscript/case-lsat-z3py-ambiguity.tex}
    \caption{Two typical types of ill-defined logical programs which can be detected during inference}
   \label{fig:case-incorrect-z3py}
\end{figure}

\zhiyu{Our framework targets first-order logic (FOL) as the reasoning formalism.} 
\ourmethodabbr{} performs \emph{inference-time} ensemble over $k$ independent \emph{AF paths}.
Each path $i$ is a complete AF pipeline: (1)
\emph{Draft} a plan $p_i$ in natural language given an input instance $x$; (2) \emph{Generate} a formalization (e.g., a Z3Py program for AR-LSAT) $z_i$ conditioned on $p_i$; (3) \emph{Repair} $z_i$ using standard solver-feedback fixes until it executes or reaches a fixed repair limit; (4) \emph{Execute} $z_i$ to derive feasible hypothesis set $S_i$; (5) \emph{Prune} ill-defined paths; and (6)  \emph{Aggregate} answers from surviving paths by majority vote.
Both plan drafting and formalization generation are performed by an LLM via fixed prompts with in-context learning (ICL).
Figure~\ref{fig:algorithm-pipeline} illustrates the pipeline.
All paths are independent samples; we do not perform tree search or branching within a path.

Our design is motivated by a common brittleness pattern we have observed in existing auto-formalization (AF) pipelines.
Even when feedback from a solver is used to repair the syntax, AF systems can fail either because the generated solver program cannot be executed due to syntactic failures, or because it can be executed but encodes incorrect semantics, a failure denoted by \emph{semantic unfaithfulness}.
To disentangle these failure modes, we follow a common practice in the AF literature and use the execution rate and the accuracy among executable formalizations as complementary diagnostics that localize errors due to executability versus incorrect semantics. The formal definitions are provided in Section~\ref{sec:metrics}. Prior AF baselines exhibit both types of failures on challenging inputs, suggesting that improving executability alone is insufficient and that searching over semantically faithful encodings remains a key bottleneck.

This diagnosis directly motivates the two components of \ourmethodabbr{}. Drafting encourages a form of controlled semantic diversity by first generating an explicit natural-language plan and then conditioning program generation on that plan.
\zhiyu{This design introduces stochasticity at the planning stage while keeping program generation deterministic via greedy decoding, which preserves syntactic correctness while increasing the chance of producing a faithful encoding.}
Pruning improves the reliability of the result by filtering executable but ill-defined programs  using solver-verifiable well-definedness checks before selecting an answer. As further discussed below, we call ill-defined a program that is contradictory or amenable to ambiguous encodings. Predictions from surviving paths are finally aggregated by majority voting.


\subsection{Drafting plans to induce semantic diversity}
\label{sec:draft}

A key challenge in AF is that many distinct symbolic encodings can appear plausible from the same natural-language input, while only a small subset is semantically faithful. Uniformly sampling programs directly from the LLM can significantly increase syntax and runtime errors. We therefore introduce stochasticity at the \emph{plan} level. 
As illustrated in the middle of Figure~\ref{fig:algorithm-pipeline}, 
a plan $p_i$ is a structured natural-language checklist that specifies:
(i) the problem entities and the variables representing their properties or assignments,
(ii) the domains of the entities and variables and the key relations among them; and 
(iii) how each textual constraint maps to solver assertions.

\zhiyu{For each path $i$, we first \emph{draft} a plan
\[
p_i \sim \pi(p \mid x),
\]
and then \emph{generate} a formalization that is compatible with the solver 
\[
z_i \sim \pi(z \mid (x, p_i)).
\]
The formalization is subject to a \emph{repair} step, if it is not executable. $\pi(\cdot)$ denotes the distribution induced by the LLM under the specified decoding strategy. In practice, a plan is sampled with temperature $T_{\textsc{draft}}$, while program generation is performed with temperature $0$ for stability.}





\subsection{Pruning paths to improve faithfulness}
\label{sec:prune}



\zhiyu{Executing a path program $z_i$ produces a feasible hypothesis set 
\[
S_i = S(z_i) \subseteq \mathcal{A},
\]
where $\mathcal{A}=\{Q_1,\ldots,Q_m\}$ denotes the set of candidate hypotheses, and $S(\cdot)$ maps $z_i$ to the set of hypotheses $S_i$ got from solver execution. Operationally, $S_i$ is obtained by issuing queries for each hypothesis $Q_j$ in the execution process.}
\zhiyu{Depending on the input context, these queries may test different logical relations, as mentioned in Section~\ref{sec:deductive-reasoning}.}


\zhiyu{However, executable programs may still encode the problem incorrectly due to semantic unfaithfulness, which is revealed during solver execution. If the solver returns no feasible hypothesis or multiple feasible hypotheses, the induced formalization violates the assumption that the task admits a single correct hypothesis. We therefore introduce well-definedness criteria to prune such paths.}

\paragraph{Well-definedness criteria.}
\zhiyu{We retain path $i$ if and only if $|S_i| = 1$.
Equivalently, we define $g_i = \mathbf{1}_{(|S_i| = 1)}$.
If $|S_i|=0$, the formalization is contradictory, producing a failing \emph{existence} check; if $|S_i|>1$, it is deemed as ambiguous, producing a failing \emph{uniqueness} check.}
In either case, exemplified in Figure~\ref{fig:case-incorrect-z3py}, the path will be pruned at inference time after verification using the solver.

\subsection{Majority voting over surviving paths}
\label{sec:agg}
Let $\mathcal{I} = \{i : g_i = 1\}$ denote the set of surviving paths. For each $i \in \mathcal{I}$, the feasible set is a singleton.
We then predict
\[
\hat{y}
=
\arg\max_{Q_j \in \mathcal{A}}
\sum_{i \in \mathcal{I}} \mathbf{1}_{(S_i = \{Q_{j}\})}.
\]

If all paths are pruned or unexecutable, the answer will be labeled as incorrect.
We aggregate over multiple sampled reasoning paths, but each path corresponds to a formalization-and-execution pipeline that could be found invalid.

\paragraph{Probabilistic interpretation.} Overall, the \ourmethodabbr{} procedure can be viewed as \textit{sampling-based marginalization} over latent plans and latent formalizations with validation gating, i.e., 
\[
\begin{aligned}
MC(Q_j \mid x)
&=
\sum_{p}\sum_{z}
\mathbf{1}_{(|S(z)| = 1, S(z)=\{Q_j\})}\ \\
& \cdot
\pi(z \mid (x,p))\,\pi(p \mid x).
\end{aligned}
\]

\noindent \ourmethodabbr{} approximates
\[
\hat{y}
\approx
\arg\max_{Q_j \in \mathcal{A}}
MC(Q_j \mid x)
\] 
by sampling $k$ independent $(p_i,z_i)$ pairs and aggregating answers from verifier-approved paths.

\newcounter{baseline}
\newcommand{\baseline}{\refstepcounter{baseline}(\thebaseline)\ }

\section{Experiments}
\label{sec:experiments}

\subsection{Datasets}

We evaluate our approach on four representative reasoning benchmarks, which jointly cover both natural and synthetic settings and assess different aspects of logical and deductive reasoning:

\noindent\textbf{AR-LSAT} is derived from analytical reasoning questions in the Law School Admission Test (LSAT)~\citep{ar-lsat}. Each instance provides a set of natural-language premises \(P\) together with multiple candidate hypotheses \(\{Q_1,\ldots,Q_m\}\). Each question specifies a target logical relation, including \emph{``must be true''} (\(P \models Q_i\)), \emph{``must be false''} (\(P \models \lnot Q_i\)), \emph{``could be true''} (\(P \land Q_i\) is satisfiable), and \emph{``could be false''} (\(P \land \lnot Q_i\) is satisfiable). This benchmark is widely regarded as challenging, with existing methods still far from perfect accuracy.
\zhiyu{We report our main results on the original public AR-LSAT test set for comparability with prior work. We additionally identified annotation issues in this benchmark and constructed a corrected 229-sample test split in Appendix~\ref{sec:appx-ar-lsat-fixed}.}

\noindent\textbf{ProofWriter} evaluates multi-step deductive reasoning over natural-language rules and facts.
Each instance consists of premises \(P\) and a single hypothesis \(Q\), which must be classified as entailed (\(P \models Q\)), contradicted (\(P \models \lnot Q\)), or unknown (neither holds)~\citep{tafjord2021proofwriter}.

\noindent\textbf{PrOntoQA} is a synthetic benchmark designed to test systematic deductive reasoning~\citep{prontoqa}.
It follows the same single-hypothesis formulation as ProofWriter, but restricts labels to binary outcomes, where each hypothesis is either entailed or contradicted by the premises.

\noindent\textbf{LogicalDeduction}, from BigBench~\citep{logicaldeduction}, also presents multiple candidate hypotheses under the same premises \(P\), but each question asks for the unique hypothesis \(Q_i\) that is logically entailed by the premises, that is, the option satisfying \(P \models Q_i\). Problems typically involve inferring object orderings or relations from a small set of constraints.

\subsection{Baselines}

To compare our method against existing LLM-based reasoning approaches, we consider a diverse set of baselines spanning prompting-based and AF-based methods:
\baseline \textbf{Direct} prompting and \textbf{CoT}~\citep{wei2022chain};
\baseline \textbf{CoT-SC} \citep{wang2022self}, an enhanced decoding approach that samples multiple reasoning paths and aggregates their outputs to improve robustness and accuracy;
\baseline \textbf{Logic-LM} \citep{logic-lm}, which first translates the natural language problem into a symbolic form and then delegates reasoning to a symbolic solver, with syntax repair based on solver feedback over the translation process;
\zhiyu{\baseline \textbf{LINC}~\citep{olausson2023linc}, 
which follows a similar way as Logic-LM but uses majority voting to aggregate multiple formalizations;}
\baseline \textbf{DetermLR} \citep{sun2024determlr}, which transitions uncertain premises into determinate conclusions through iterative control and, potentially, external modules (e.g., memory);
\baseline \textbf{SymbCoT} \citep{symbcot}, which injects symbolic reasoning operations into the chain-of-thought framework;
\baseline \textbf{Logic-LM++}~\citep{logic-lm++}, which extends Logic-LM by introducing multi-step refinement and pairwise comparisons;
\baseline \textbf{CLOVER}~\citep{clover}, which introduces a compositional translation of natural language into first-order logic via logical dependency structures, coupled with verification through symbolic solvers to ensure semantic correctness.
\zhiyu{\baseline \textbf{MAD-LOGIC}~\citep{yang2026enhancing}, a multi-agent debate framework that improves symbolic translation and reasoning through debate among symbolic and natural-language agents, with final answers selected by majority vote.}


\begin{table*}[h!]
\caption{End-to-end accuracy (\Acc{}) on four deductive reasoning benchmarks. We compare prompting-based methods and auto-formalization (AF) methods, without fallback (abstention) and with CoT fallback. \zhiyu{``--'' indicates that the corresponding setting is not supported by the baseline.}} 
\label{tab:main-results}

\centering

\resizebox{\linewidth}{!}{

\begin{tabular}{ll|c|c cccc}
\toprule

\multirow{2}{*}{Paradigm} &
\multicolumn{1}{l}{\multirow{2}{*}{Method}} &
\multicolumn{1}{c}{ {Fallback} } &
\multirow{2}{*}{LLM} &
\multicolumn{4}{c}{End-to-end Accuracy (\%)} \\


& 
\multicolumn{1}{l}{} &
\multicolumn{1}{c}{ {Policy} } &
& {AR-LSAT} & {ProofWriter} & {PrOntoQA} & {LogicalDeduction} \\

\midrule\midrule

\multirow{8}{*}{Prompting} 
    & Direct                & \multirow{8}{*}{N/A} & \multirow{5}{*}{GPT-4}
                              & 33.30   & 52.67     & 77.40     & 71.33 \\
    & CoT                   & & & 35.06   & 68.11     & 91.00     & 74.67 \\
    & CoT-SC                & & & --      & 69.33     & 93.40     & 74.67 \\ 
    & DetermLR              & & & --      & 79.17     & 98.60     & 85.00 \\
    & SymbCoT               & & & 43.91   & 82.50     & 99.60     & 93.00 \\
    \cline{2-2}\cline{4-8}
    & Direct           &  & \multirow{4}{*}{GPT-4o}  & 30.30 & 53.70 & -- & 84.70 \\ 
    & CoT           &  & & 36.09 & 72.67  &   99.00   &  91.33 \\ 
    & CoT-SC           &  & & 38.37 & --  &   --   &  -- \\ 
    & SymbCoT   &   &  & 34.20 &  80.20 & -- &  90.70    \\ 

\midrule

\multirow{10}{*}{AF}
    & Logic-LM                              & \multirow{5}{*}{\makecell{Abstention \\ (None)}} & \multirow{5}{*}{GPT-4}      & 19.56 & 79.00 & 83.00  & 88.00 \\
    & Logic-LM++                            & &       & 21.18 & 78.80 &  --   &  --   \\
    & LINC & & & 37.83 & 98.30 & -- & -- \\
    & MAD-LOGIC & &  & 53.25 & 92.00 & 100.00 & 94.33 \\
    & \textbf{\ourmethodabbr{}}             & &       & 78.43  & 98.33 & 99.32 & 99.50 \\
    \cline{2-2}\cline{4-8}
    & CLOVER                                &  & \multirow{3}{*}{GPT-4o}  & 46.80 &   96.50& --        & 99.00  \\
    & LINC & & & 44.65 & 99.50 & -- & -- \\
    & \textbf{\ourmethodabbr{}}             &   &  & 78.00    & 99.67 & 100.00    & 100.00 \\
    
    \cline{2-8}
    
    & Logic-LM                  & \multirow{5}{*}{CoT} &  \multirow{3}{*}{GPT-4}     & 43.04 & 79.66 & 83.20 & 87.63 \\
    & Logic-LM++                &  &      & 46.32 & 79.66 & --    & --    \\
    & \textbf{\ourmethodabbr{}} & &     & 80.65  & 98.33 & 99.36 & 99.50  \\
    \cline{2-2}\cline{4-8}
    & CLOVER                     & &
    \multirow{2}{*}{GPT-4o} & 62.80     & 96.70 & --        &  99.30 \\
    & \textbf{\ourmethodabbr{}} & &      & 82.83 & 99.67 & 100.00 & 100.00 \\

\bottomrule
\end{tabular}

}

\end{table*}

\subsection{Evaluation Metrics}
\label{sec:metrics}

We report end-to-end \textbf{accuracy} (\Acc{}), defined as the fraction of instances whose \emph{final} predicted answer matches the ground truth.
Let $N$ be the number of instances, and let $c_j\in\{0,1\}$ indicate whether the final output for instance $j$ is correct.
\begin{align}
\Acc{} & = \frac{1}{N}\sum_{j=1}^N c_j
\label{eq:acc}
\end{align}
To diagnose brittleness of AF-only methods (i.e., without fallback via prompting), we additionally report \textbf{execution rate} (\ExecRate{}) and \textbf{accuracy among executable formalizations} (\AccAtExec{}).
For an AF-only method, the system attempts to generate solver programs and derive an answer through execution and, when applicable, pruning.
\zhiyu{If no \emph{valid} AF path is available for an instance, the system abstains and the instance is counted as incorrect.}

\begin{align}
\ExecRate{}  &= \frac{1}{N}\sum_{j=1}^N e_j, \label{eq:exec}
\end{align}
\begin{align}
\AccAtExec{} &= \frac{\sum_{j=1}^N e_j c^{\mathrm{AF}}_j}{\sum_{j=1}^N e_j}.
\label{eq:accatexec}
\end{align}

\zhiyu{Let $e_j\in\{0,1\}$ indicate whether the AF procedure for instance $j$ yields at least one \emph{valid} AF path.
A path is considered \emph{valid} if its solver program executes successfully (i.e., without errors/exceptions within a fixed timeout) and, when pruning is enabled, passes the pruning criteria.
Let $c^{\mathrm{AF}}_j\in\{0,1\}$ indicate whether the final AF prediction for instance $j$ is correct, defined only when $e_j=1$.}

For hybrid systems (e.g., Logic-LM with CoT fallback), the final answer may come from either the AF component or a fallback when AF fails; we report end-to-end accuracy without diagnostics.

\subsection{Experimental Setup}

\paragraph{Experimental setup.}
We conduct our experiments using GPT-4 and GPT-4o, consistent with baselines.
For all LLM API calls, the maximum output length is set to 2,048 tokens, with nucleus sampling parameter $p=1.0$. 
Plan drafting uses temperature $T_{\textsc{draft}}=1.0$,
while program generation uses greedy decoding ($T_{\textsc{gen}}=0$).
Each path allows at most $R=2$ rounds of solver-feedback repair.
Plan and program generation use three-shot in-context learning (ICL) from training data, while syntax repair is zero-shot. Prompt templates are provided in Appendix~\ref{sec:prompt-schema}.
If there is a tie (no strict majority), we choose the first-appearing answer for determinism.
All experiments are repeated over 10 runs. We report mean (and standard deviation when applicable).

\subsection{Main Results}
\label{sec:main-results}

Table~\ref{tab:main-results} reports end-to-end accuracy (\Acc{}) on four deductive reasoning benchmarks, covering both prompting-based methods and auto-formalization (AF) methods. For AF, we evaluate two settings: \emph{AF-only} (no fallback) and \emph{AF with CoT fallback}, where CoT is used only when AF fails. This separation allows us to distinguish improvements to the AF itself from gains due to CoT fallbacks.

\zhiyu{We first compare under the AF-only setting. On AR-LSAT, the strongest AF baselines in our table are MAD-LOGIC (GPT-4), which achieves 53.25\%, and CLOVER (GPT-4o), which achieves 46.8\%. 
In contrast, \ourmethodabbr{} reaches 78.43\% with GPT-4 and 78.00\% with GPT-4o, improving over the strongest baselines by 25.2 and 31.2 absolute points, respectively. 
Notably, MAD-LOGIC relies on a multi-agent debate framework, whereas \ourmethodabbr{} uses a single-model inference-time pipeline.}

AF baselines also use CoT fallbacks for failed formalization. To make fair comparisons, we also report in the same setting. With this setting, \ourmethodabbr{} remains strong: on AR-LSAT, it achieves 80.65\% (GPT-4) and \AccOnLSATOurGPTFourO{} (GPT-4o), and it still saturates ProofWriter, PrOntoQA, and LogicalDeduction. 

Finally, we compare to prompting-based methods. Among the baselines where results are available, SymbCoT is typically the strongest. On AR-LSAT, it achieves 43.91\% with GPT-4 and 34.20\% with GPT-4o. Meanwhile, \ourmethodabbr{} attains 78.43\% (AF-only) and 80.65\% (with CoT fallback) under GPT-4, and \AccAFOnLSATOurGPTFourO{} (AF-only) and \AccOnLSATOurGPTFourO{} (with CoT fallback) under GPT-4o.
\zhiyu{
Additional experiments on newer model families such as GPT-5.2~\citep{openai2025gpt52} and Gemini-3.0~\citep{gemini3} are reported in Appendix~\ref{sec-pruning-on-new-models}--\ref{sec:generalization}.
}

\subsection{In-depth Analysis}

\begin{table}[h]
\caption{Decomposed AF diagnostic metrics on AR-LSAT. $k$ denotes the number of sampled AF paths.}
\label{tab:af-diagnostics-metrics}
\centering
\resizebox{\linewidth}{!}{

\begin{tabular}{l l c c c }
\toprule 
Method & LLM
& \makecell{\Acc{} (\%)}
& \makecell{ $\mathbf{Exec}$ \\ $\mathbf{Rate}$ (\%)}
& \makecell{\Acc{}(\%)\\$\mathbf{@Exec}$} \\
\midrule\midrule

Logic-LM & \multirow{3}{*}{GPT-4}
&  19.6   &       32.6     &   60.0          \\


\ourmethodabbr{} ($k=1$) & &
36.4 &   43.3    &     84.1        \\

\ourmethodabbr{} ($k=20$) & &
78.4 &   91.7    &      85.5       \\

\midrule

CLOVER   & \multirow{3}{*}{GPT-4o} &
46.8    &      59.7        &       78.3      \\



\ourmethodabbr{} ($k=1$) & &
39.9    &    44.6     &     89.4         \\

\ourmethodabbr{} ($k=20$) & &
78.0    &    89.6     &     87.1         \\

\bottomrule
\end{tabular}

}
\end{table}

Table~\ref{tab:af-diagnostics-metrics} shows that \ourmethodabbr{} surpasses the baselines in both \ExecRate{} and \AccAtExec{} when using a larger number of sampled paths ($k{=}20$).
At $k{=}1$, \ourmethodabbr{} trails CLOVER in \ExecRate{} because limited sampling reduces the chance of producing a valid formalization that survives pruning.

To test our hypothesis that a few AF attempts are unlikely to yield a faithful formalization on hard problems, we examine how performance changes as we sample more candidate paths.
We define a diagnostic metric, $\mathrm{cover}@k$, which marks an instance as solved if \emph{any} of the $k$ sampled AF paths yields the correct answer.
This metric characterizes the \emph{potential} of AF under broader exploration: higher $\mathrm{cover}@k$ indicates a greater likelihood that at least one correct formalization appears among the $k$ sampled paths.

\begin{figure}[h]
    \centering
    \includegraphics[width=0.7\linewidth]{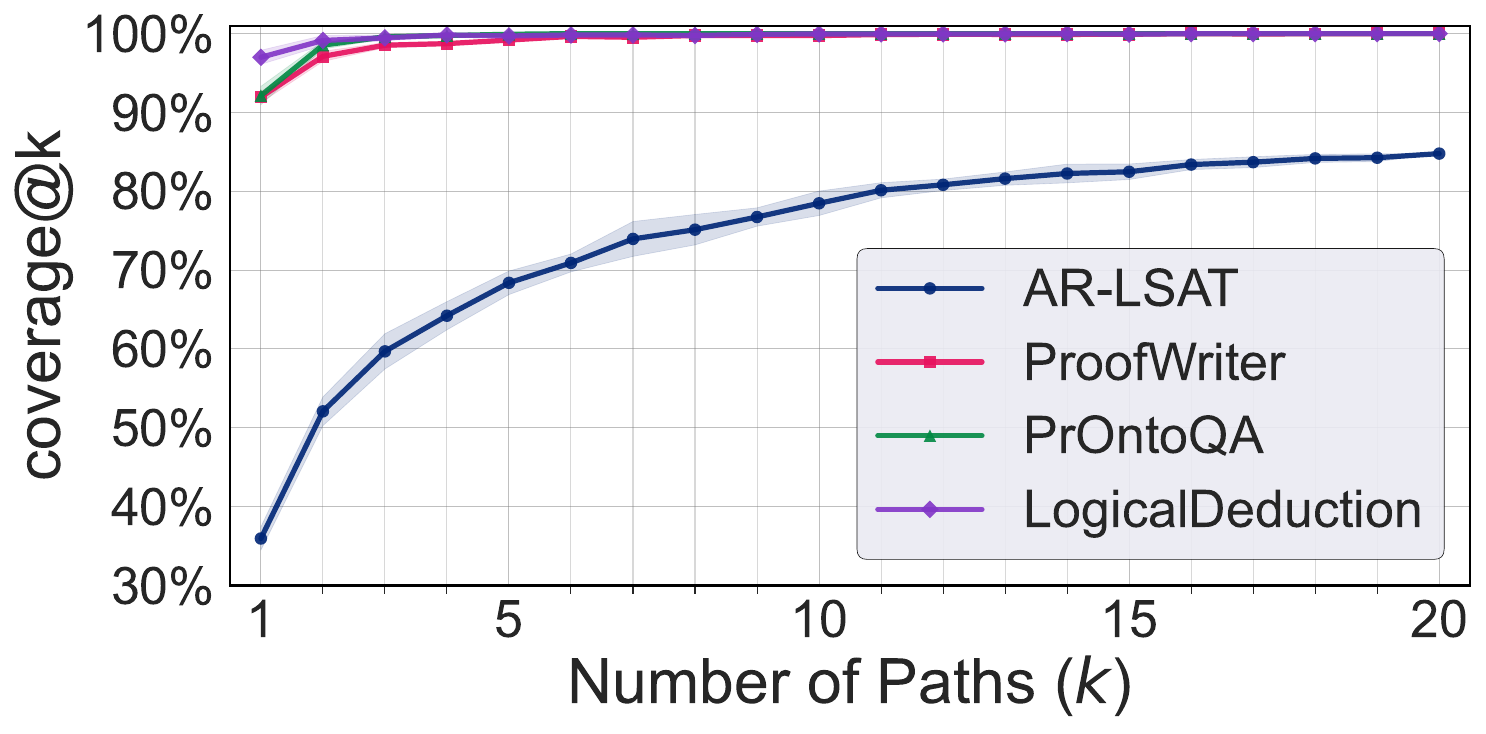}
    \caption{Coverage ($\mathrm{cover}@k$) as a function of the number of sampled AF paths $k$ across four benchmarks. 
    }
    \label{fig:coverage-vs-k}
\end{figure}

Figure~\ref{fig:coverage-vs-k} shows that $\mathrm{cover}@k$ increases monotonically with $k$, indicating that the chance of obtaining a correct answer grows as we explore more candidates.
Moreover, the gap between $\mathrm{cover}@k$ and realized AF accuracy at small $k$ suggests that many failures are due to under-exploration of the formalization space rather than the absence of faithful formalizations.
In other words, many instances are solvable by AF, but not reliably within a single or a few attempts.
\zhiyu{Cost-normalized comparisons with CoT-SC are reported in Appendix~\ref{sec:cost-normalized}.}

\subsection{Effectiveness of drafting plans}
\label{sec:ablation-draft}

\begin{table}[h!]
\caption{Accuracy with a naive AF path w/ and w/o drafted plans \zhiyu{(on GPT-4, w/o pruning).}}
\label{tab:ablation-study-draft}
\centering
\resizebox{\linewidth}{!}{
\begin{tabular}{l  c c}
\toprule
& \multicolumn{2}{c}{\AccAF{} (\%) } \\

                    & AF w/o \VerbGen{}              & AF w/ \VerbGen{}     \\
\midrule\midrule
AR-LSAT             & 36.35 $\pm$ 1.22 & 41.78 $\pm$ 2.46 \\
ProofWriter         &       97.69 $\pm$ 0.18           &   98.33 $\pm$ 0.22               \\
PrOntoQA            &       94.92 $\pm$ 0.84           &       99.56 $\pm$ 0.18           \\
LogicalDeduction   &            98.60 $\pm$ 0.58      &     98.23 $\pm$ 0.32             \\
\bottomrule
\end{tabular}
}
\end{table}

Table~\ref{tab:ablation-study-draft} studies the effect of drafting a plan before formalization generation in a single-path AF setting without pruning.
On AR-LSAT, adding drafted plans improves \AccAF{} by over 5\%, indicating that explicit planning helps AF improve performance even without path selection or aggregation.
On PrOntoQA, although plain AF already achieves high accuracy, drafted plans further reduce occasional failures and push performance closer to saturation.
On ProofWriter and LogicalDeduction, both variants achieve near-ceiling performance, and drafting plans neither degrades accuracy nor introduces instability.



\subsection{Sensitivity to Number of Paths}

    

\begin{figure}[t]
    \centering
    \begin{subfigure}[b]{0.48\linewidth}
        \centering
        \includegraphics[width=\textwidth]{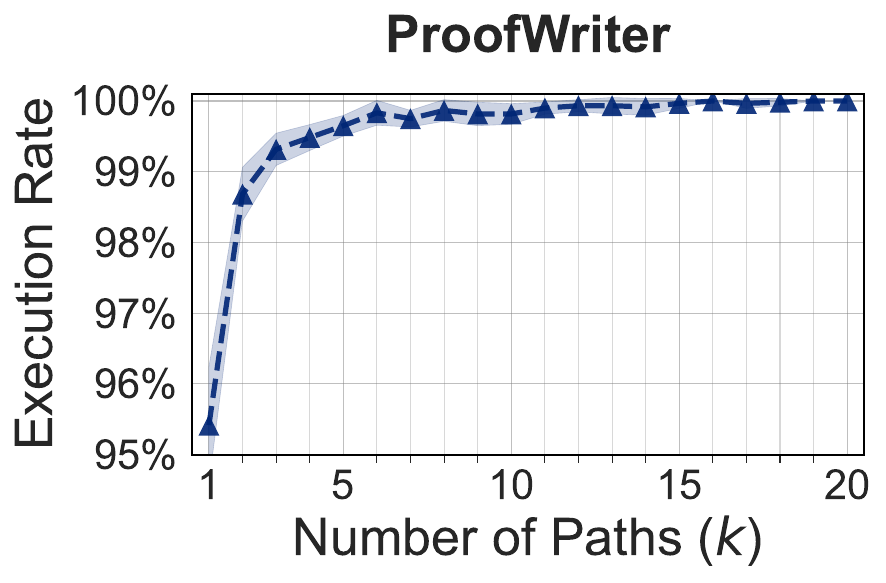}
        \caption{ExecRate}
        \label{fig:proofwriter-exec-rate}
    \end{subfigure}
    \hfill
    \begin{subfigure}[b]{0.48\linewidth}
        \centering
        \includegraphics[width=\textwidth]{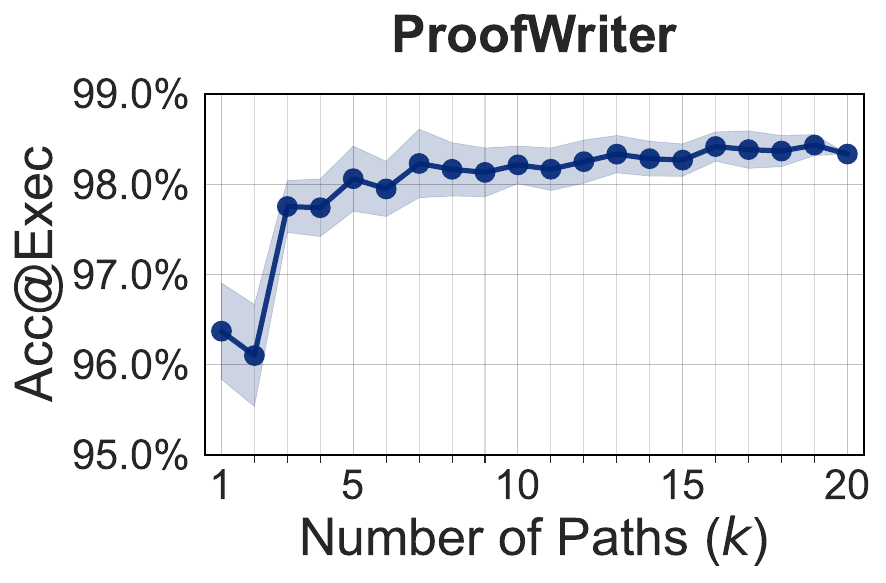}
        \caption{Acc@Exec}
        \label{fig:proofwriter-exec-acc}
    \end{subfigure}
    \caption{Diagnostic metrics on ProofWriter across different numbers of paths ($k$).}
    \label{fig:proofwriter-exec-metrics-and-k}
\end{figure}

We first study the effect of sampling multiple AF paths with drafted plans on ProofWriter. As shown in Figure~\ref{fig:proofwriter-exec-metrics-and-k}, increasing the number of paths $k$ leads to a rapid increase in both \ExecRate{} and \AccAtExec{}. This behavior indicates that, when semantic ambiguity is limited, drafted plans combined with multi-path aggregation can reliably increase AF accuracy without introducing instability.


\begin{figure}[h]
    \centering
    \includegraphics[width=0.5\linewidth]{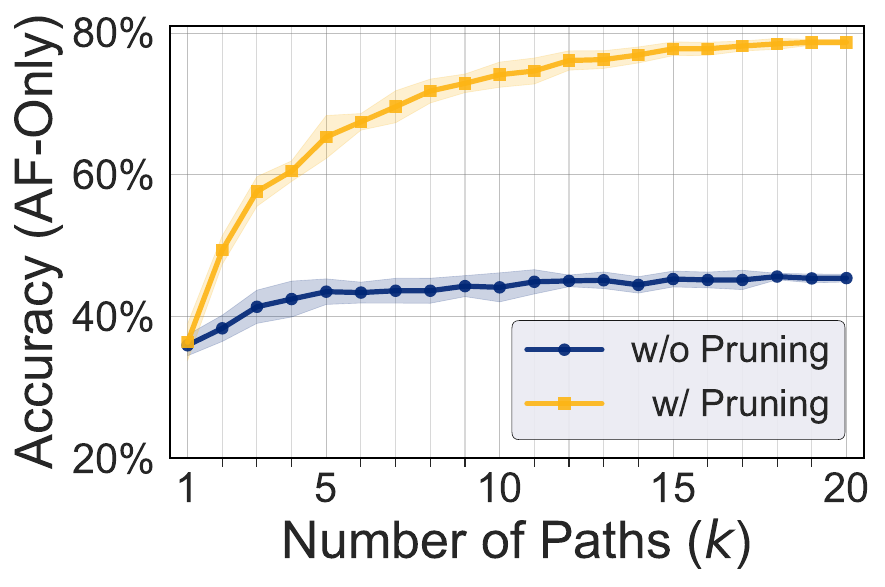}
    \caption{\AccAF{} on AR-LSAT w/o vs. w/ pruning across different numbers of paths ($k$). Without pruning, the \AccAF{} plateaus around $45\%$ quickly, while pruning enables further accuracy boost to around $78\%$.}
    \label{fig:ablation-ensemble}
\end{figure}

We examine on more challenging benchmarks like AR-LSAT. Figure~\ref{fig:ablation-ensemble} shows that accuracy on AR-LSAT increases almost monotonically with larger $k$. Without pruning, accuracy improves by about 9\% when increasing the number of paths from $k=1$ to $k=10$, but gains less than 1\% when further increasing $k$ to $20$.

This plateau suggests that simply increasing the number of AF paths is insufficient. Many additional paths are executable but semantically ill-defined, producing contradictory or answer-ambiguous paths that cannot be resolved by aggregation alone. This observation motivates a closer examination of pruning.




\subsection{Effectiveness of Pruning}
\label{sec:ablation-prune}
Table~\ref{tab:ablation-pruning-overall} shows that well-definedness pruning is critical on AR-LSAT, improving \AccAF{} from $45.13\%$ to $78.43\%$ (+33.30\%).
In contrast, pruning has negligible effect on ProofWriter, PrOntoQA, and LogicalDeduction, where accuracy is already near ceiling, and differences are within variation.
Overall, pruning mainly helps harder problems by filtering ill-defined AF candidates before aggregation, while remaining safe on easier benchmarks.

\begin{table}[h]
\caption{The \AccAF{} (\%) of \ourmethodabbr{} w/o vs. w/ pruning on four benchmarks. There are twenty paths ($k = 20$), and every path uses drafted plans.}
\label{tab:ablation-pruning-overall}
\centering
\resizebox{\linewidth}{!}{

\begin{tabular}{l c c}
    \toprule
    & \multicolumn{2}{c}{\AccAF{}(\%)} \\
        &  w/o Pruning & w/ Pruning  \\
    \midrule\midrule
    AR-LSAT  & 45.13 $\pm$ 0.34 & 78.43 $\pm$ 0.55  \\
    ProofWriter & 98.33 $\pm$ 0.00 & 98.33 $\pm$ 0.00 \\
    PrOntoQA & 99.34 $\pm$ 0.10 & 99.32 $\pm$ 0.10  \\
    LogicalDeduction & 99.33 $\pm$ 0.00 & 99.50 $\pm$ 0.18  \\
    \bottomrule
\end{tabular}

}
\end{table}


\begin{figure}[htbp]
    \centering
    \begin{subfigure}[b]{0.48\linewidth}
        \centering
        \includegraphics[width=\linewidth]{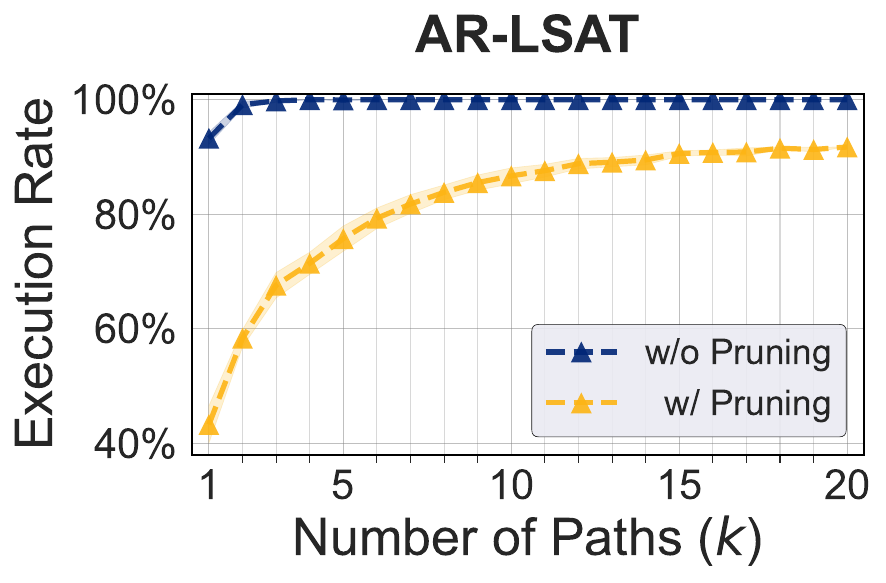}
        \caption{\ExecRate{}}
        \label{fig:ar-lsat-exec-rate}
    \end{subfigure}
    \hfill
    \begin{subfigure}[b]{0.48\linewidth}
        \centering
        \includegraphics[width=\linewidth]{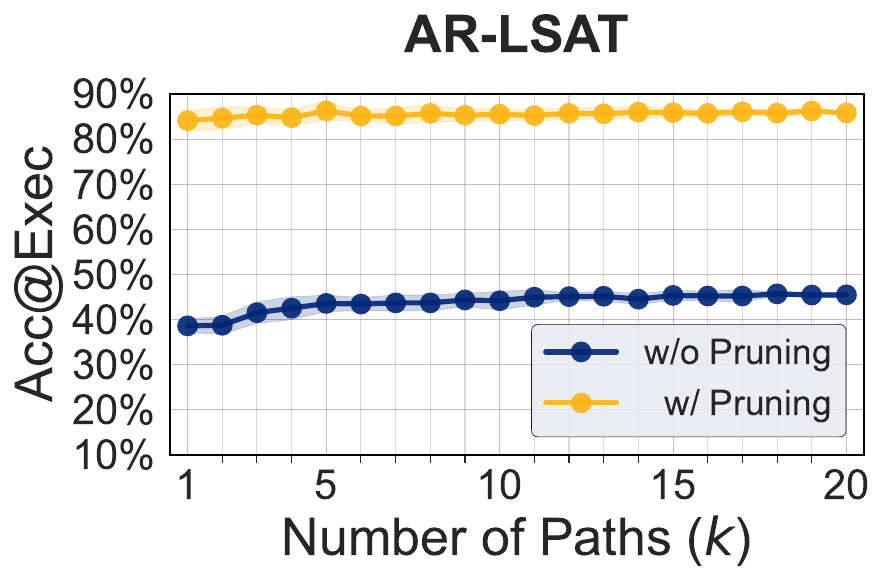}
        \caption{\AccAtExec{}}
        \label{fig:ar-lsat-exec-acc}
    \end{subfigure}
    \caption{Diagnostic metrics on AR-LSAT w/o vs. w/ pruning across different numbers of paths ($k$).}
    \label{fig:ar-lsat-pruning-comparison}
\end{figure}

Figure~\ref{fig:ar-lsat-pruning-comparison} further clarifies this effect using diagnostic metrics on AR-LSAT.
As shown in Figure~\ref{fig:ar-lsat-exec-rate}, pruning substantially reduces \ExecRate{}, since many sampled paths are discarded because they are contradictory or answer-ambiguous.
However, Figure~\ref{fig:ar-lsat-exec-acc} shows that pruning yields a large and consistent increase in \AccAtExec{} across all values of $k$.
This suggests that pruning removes semantically ill-defined formalizations while retaining paths that are more likely to be faithful.
Consequently, although fewer paths survive, the remaining ones are substantially more reliable, which improves overall \AccAF{} after aggregation. \zhiyu{A detailed failure analysis is provided in Appendix~\ref{sec:failure-analysis}.}

\section{Conclusion}




We present \ourmethod{}, an inference-time framework that improves auto-formalization via plan-conditioned generation, solver-based pruning, and aggregation. Across four benchmarks, it improves end-to-end accuracy and reduces reliance on fallback prompting. Our analysis suggests that, once executability is addressed, performance is mainly limited by searching for semantically correct formalizations, motivating stronger validation and more efficient exploration.

\section{Limitations}
\label{sec:limitations}

Our work improves AF through inference-time ensembling, verification-guided pruning, and aggregation. We note several limitations.

\paragraph{Inference-time cost.}
Our approach samples multiple candidate formalizations and executes solver checks, which increases inference-time compute, latency, and API cost relative to single-pass AF. While it is lightweight, the overall system cost grows with the number of candidates and solver invocations. How to achieve an accuracy-cost trade-off under strict budgets remains important.

\paragraph{No adaptation of the base generator.}
We treat the foundation model as a black box and do not improve the AF generator via fine-tuning, reinforcement learning, or distillation. As a result, systematic translation errors of the base model may persist. Combining our framework with model adaptation is a promising direction.


\paragraph{Aggregation with equivalence awareness.}
We aggregate predictions at the answer level and do not cluster or de-duplicate candidates by symbolic equivalence. This may overweight redundant formalizations and miss opportunities for more principled equivalence-aware aggregation. Developing equivalence-aware or diversity-aware aggregation is an interesting direction for future work.


\section{Ethical Considerations}

This work studies auto-formalization for deductive reasoning on standard academic benchmarks that contain no personal, sensitive, or user-generated data, so privacy and consent issues do not arise. A key risk of auto-formalization is producing executable but semantically incorrect programs, which could be harmful if applied in high-stakes domains; our method mitigates this risk by using solver-verified checks to remove contradictory or answer-ambiguous formalizations before aggregation, though it does not guarantee full semantic correctness. The approach relies on external large language models and therefore inherits their limitations and biases, and it increases inference-time cost due to multi-path sampling and solver calls. As a result, the method should be viewed as a research system for controlled settings rather than a drop-in replacement for human-verified reasoning in real-world applications.

\bibliography{arxiv-references}

\appendix

\section{Appendix}
\label{sec:appendix}


\subsection{Propositional and First-Order Logic}
\label{appx-fol}

Propositional logic represents knowledge using \emph{atomic statements} (propositions), which are the simplest units of truth---statements that can be either true or false, such as ``it is raining'' or ``the light is on.'' These statements can be combined with logical connectives such as ``and,'' ``or,'' ``not,'' and ``if-then'' to form more complex expressions. While powerful in its simplicity, propositional logic is limited in scope: it treats entire statements as indivisible units and cannot represent relationships between objects or quantify over them~\cite{kneale1984development}. 

First-Order Logic (FOL)~\cite{barwise1977introduction} extends propositional logic with two key features: \emph{predicates}, which describe properties of objects and relations among them (e.g., $\mathit{Loves}(x, y)$: ``$x$ loves $y$''), and \emph{quantifiers}, which enable statements about all or some objects in a domain (e.g., $\forall x \, \exists y \; \mathit{Loves}(x, y)$: ``every person loves someone''; $\exists y \, \forall x \; \mathit{Loves}(x, y)$: ``someone is loved by everyone'').
While propositional logic is best for straightforward true/false combinations, First-Order Logic is much more expressive and can model complex systems involving multiple entities and their relationships~\cite{barwise1977introduction, fitting2012first}. FOL's clear syntax and powerful expressiveness underpin formal verification by enabling precise definitions and guaranteed logical inference~\cite{fitting2012first}.

\subsection{NL to First-Order Logic}
\label{appx-nl2fol}
Recent papers~\cite{yang2023harnessing, lalwani2024autoformalizing} propose using LLMs to translate natural language reasoning problems into First-Order Logic (FOL).

\paragraph{Logical modeling in AF.}
\zhiyu{The AF process introduces a number of entities, representing objects in a problem (e.g., people, items, or time slots). A set of  
variables is used to represent properties or relations of these entities, and each variable ranges over a domain of possible values (e.g., Boolean, integers, or symbolic identifiers). 
Natural-language constraints are then translated into logical constraints over these variables and implemented as solver assertions in the generated program.}

Translation begins by specifying the domain of discourse, the set of objects under consideration (e.g., persons, numbers, graph nodes), and any basic assumptions about them. 
Next, one chooses a predicate vocabulary to describe properties and relations, along with symbols for specific objects and functions.


The process varies slightly across different benchmarks. In AR-LSAT~\cite{ar-lsat}, natural language arguments are rephrased into structured predicates that capture assumptions and logical consequences. In ProofWriter~\cite{tafjord2021proofwriter}, Stepwise inference is expressed with explicit predicates, quantifiers, and logical rules, allowing derivations to be precisely traced from facts. In PrOntoQA~\cite{prontoqa}, questions that involve multi-hop reasoning over short statements are mapped into predicates and quantifiers that explicitly represent each reasoning step. Finally, in LogicalDeduction~\cite{logicaldeduction}, the focus is on converting abstract puzzles or conditional rules into quantified statements that ensure correct logical inference.

Across these tasks, the central challenge is faithfully capturing natural-language semantics, particularly ambiguity of reference, quantifier scope, negation, modality, and implicit background knowledge, within FOL's framework.


\zhiyu{




\subsection{Auto-formalization in interactive theorem provers}
\label{sec:related-work}

A parallel line of work studies translating informal mathematical reasoning into formal proofs within interactive theorem provers (ITPs). Draft, Sketch, and Prove (DSP)~\cite{jiang2023draft} guides theorem provers such as Isabelle by first generating informal proof sketches and then refining them into formal proofs. TheoremLlama~\cite{wang2024theoremllama} fine-tunes large language models to generate Lean4 proofs directly.

These works operate in a substantially different setting from ours. DSP and TheoremLlama target full proof synthesis within higher-order logic proof assistants, where correctness is verified through interactive proof checking.

In contrast, our setting focuses on closed-end logical reasoning tasks with a determinate answer, where correctness is validated via solver execution. Methodologically, DSP follows a single-path refinement loop within an ITP environment, whereas our framework performs independent multi-path drafting followed by solver-based pruning and answer aggregation. Furthermore, TheoremLlama relies on supervised fine-tuning for Lean4 proof generation, while our approach operates purely at inference time.
\subsection{Pruning Effects on Newer Model Families}
\label{sec-pruning-on-new-models}

Figure~\ref{fig:new-models-pruning-curves} compares AF-only accuracy as the path budget $k$ increases for GPT-5.2 and Gemini-3-flash under D\&P with and without pruning. In both cases, pruning improves the efficiency-accuracy tradeoff: the pruned variant rises more quickly at small $k$ and remains consistently above the unpruned variant across the path budget range.

\begin{figure}[htbp]
\centering
\begin{subfigure}{0.48\linewidth}
    \centering
    \includegraphics[width=\linewidth]{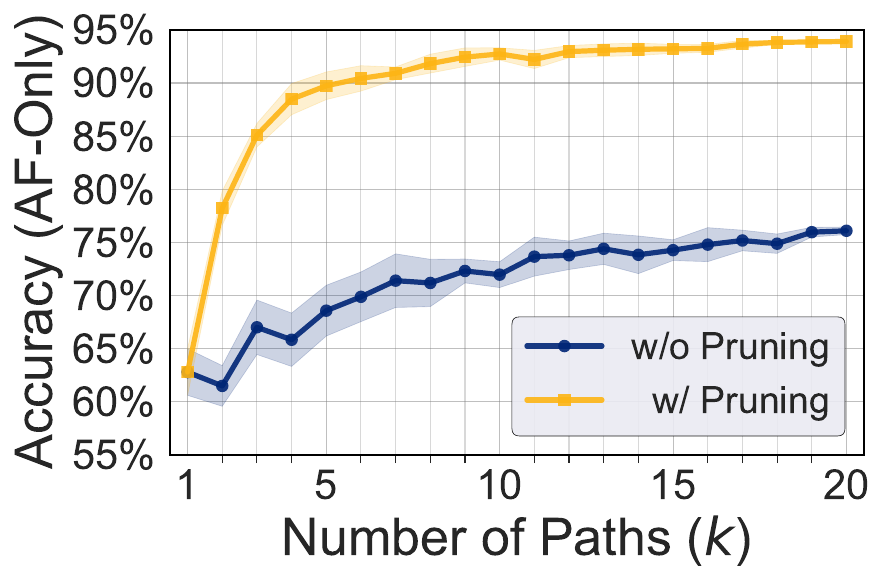}
    \caption{GPT-5.2}
\end{subfigure}
~
\begin{subfigure}{0.48\linewidth}
    \centering
    \includegraphics[width=\linewidth]{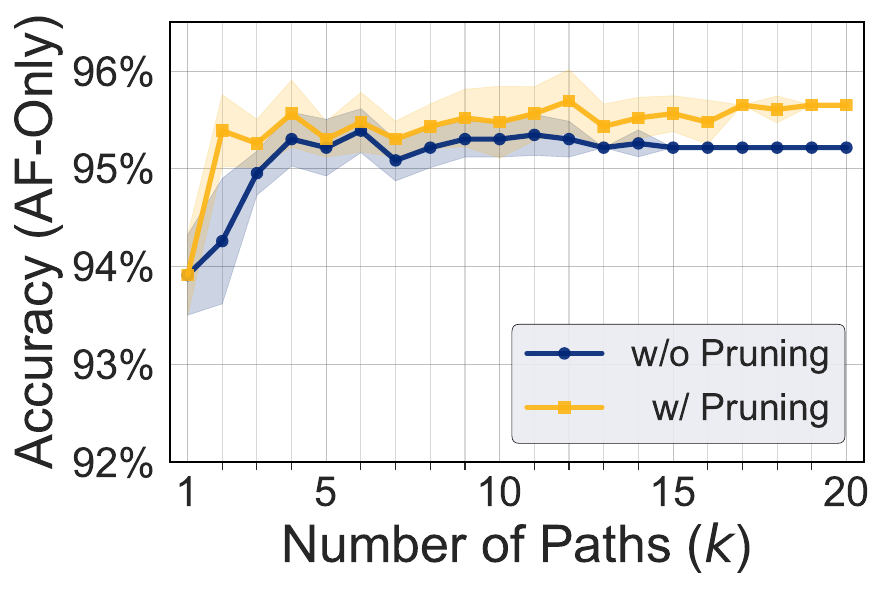}
    \caption{Gemini-3-flash}
\end{subfigure}
\caption{AF-only accuracy versus $k$ for D\&P without pruning and D\&P with pruning on AR-LSAT.}
\label{fig:new-models-pruning-curves}
\end{figure}


\subsection{CoT-SC vs.\ D\&P on Newer Model Families}
\label{sec:new-models-cotsc}

Figure~\ref{fig:new-models-cotsc-curves} compares CoT self-consistency (CoT-SC) and D\&P as the path budget $k$ increases for GPT-5.2 and Gemini-3-flash on AR-LSAT. The trend is consistent across both model families. CoT-SC improves as more paths are aggregated, but D\&P reaches a substantially higher accuracy at the same path budget. This indicates that the gains from D\&P do not come only from sampling multiple attempts; they come from combining multi-path search with explicit formalization and pruning.

\begin{figure}[htbp]
\centering
\begin{subfigure}{0.48\linewidth}
    \centering
    \includegraphics[width=\linewidth]{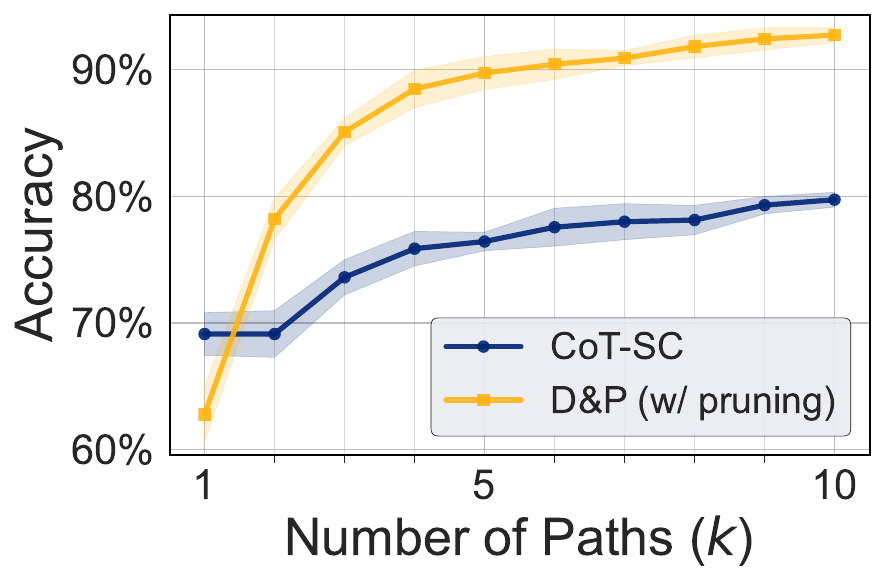}
    \caption{GPT-5.2}
\end{subfigure}
~
\begin{subfigure}{0.48\linewidth}
    \centering
    \includegraphics[width=\linewidth]{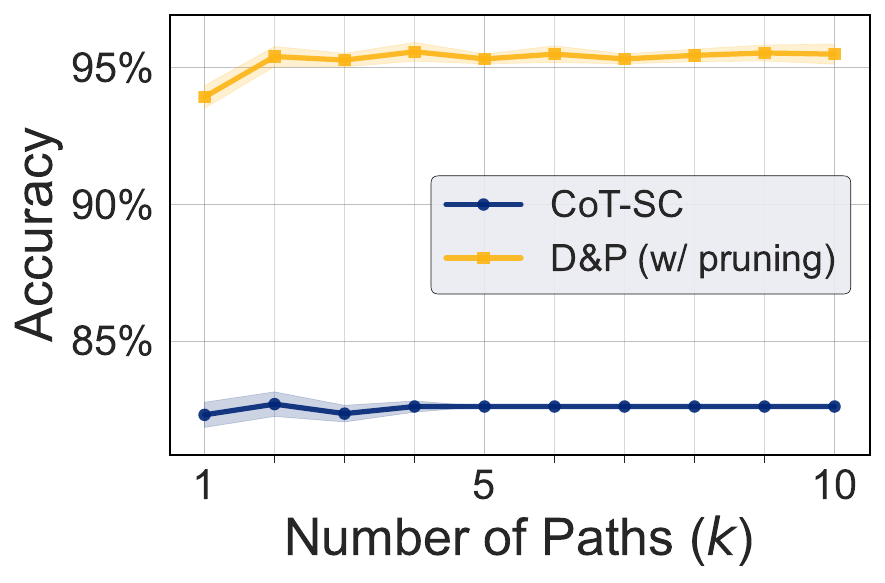}
    \caption{Gemini-3-flash}
\end{subfigure}
\caption{Accuracy versus path budget $k$ for CoT-SC and D\&P on on AR-LSAT.}
\label{fig:new-models-cotsc-curves}
\end{figure}
\subsection{Generalization Across Model Families}
\label{sec:generalization}

\begin{table}[htbp]
\centering
\small
\caption{Generalization to additional backbones on AR-LSAT ($k=10$ used for D\&P, and 3-shot used for CoT).}
\label{tab:results-other-models}
\resizebox{\columnwidth}{!}{
\begin{tabular}{l|lcccc}
\hline
Model & Method & Acc (\%) & ExecRate (\%) & Acc@Exec (\%) & Coverage@k (\%) \\
\hline
\hline
DeepSeek-R1 & CoT & 83.04 & 100.00 & 83.04 & 83.04 \\
DeepSeek-R1 & D\&P & 93.83 & 99.52 & 94.28 & 95.96 \\
\hline
GLM-4-plus & CoT & 49.13 & 100.00 & 49.13 & 49.13 \\
GLM-4-plus & D\&P & 80.00 & 93.04 & 85.98 & 86.96 \\
\hline
Kimi-K2.5 & CoT & 95.65 & 100.00 & 95.65 & 95.65 \\
Kimi-K2.5 & D\&P & 95.43 & 99.13 & 96.27 & 95.65 \\
\hline
\end{tabular}
}
\end{table}

We further evaluate D\&P on DeepSeek-R1~\citep{guo2025deepseek}, GLM-4-plus, and Kimi-K2.5~\citep{team2026kimi} shown in Table~\ref{tab:results-other-models}.

D\&P provides substantial improvements on DeepSeek-R1 and GLM-4-plus. On Kimi-K2.5, 3-shot CoT already achieves near-ceiling performance (95.65\%), leaving limited headroom for further gains. In this regime, D\&P maintains comparable accuracy while preserving solver-based execution validation.

Overall, these results suggest that the benefits of draft-and-prune are not restricted to a specific model family. More broadly, AF-based reasoning and token-based reasoning represent complementary directions, and improving AF robustness makes it competitive with strong reasoning models while enabling solver-based execution validation.

\subsection{Comparison with LINC-Style Aggregation}
\label{sec:linc-comparison}

LINC~\cite{olausson2023linc} combines LLM-based formalization with majority voting over solver-validated outputs and achieves near-ceiling performance on ProofWriter. We will clarify the relationship between LINC and D\&P below.


\paragraph{LINC-style proxy on AR-LSAT.}
Since the released LINC codebase does not include AR-LSAT, we construct a LINC-style proxy that performs majority voting on multi-paths. Table~\ref{tab:linc-arlsat} reports results under both low and high temperature settings.

\begin{table}[htbp]
\centering
\small
\caption{LINC-style proxy vs.\ D\&P on AR-LSAT (GPT-4o).}
\label{tab:linc-arlsat}
\begin{tabular}{c|ccc}
\hline
$k$ & LINC ($T=0$) & LINC ($T=1$) & D\&P \\
\hline
1  & 38.22 & 38.17 & 39.87 \\
5  & 39.30 & 44.09 & 64.04 \\
10 & 39.65 & 44.65 & 72.17 \\
15 & 40.30 & 44.52 & 75.91 \\
20 & 40.61 & 44.57 & 78.00 \\
\hline
\end{tabular}
\end{table}

The LINC-style proxy quickly plateaus around 40--45\%, whereas D\&P continues to improve as $k$ increases.

\paragraph{Fidelity validation on ProofWriter.}
To validate that our proxy faithfully reproduces LINC-style behavior, we evaluate it on ProofWriter in Table~\ref{tab:linc-proofwriter}. Original LINC paper reports near-ceiling performance on this benchmark.

\begin{table}[htbp]
\centering
\small
\caption{LINC-style proxy vs.\ D\&P on ProofWriter (GPT-4o).}
\label{tab:linc-proofwriter}
\begin{tabular}{c|cc}
\hline
$k$ & LINC & D\&P \\
\hline
1  & 98.92 & 98.60 \\
5  & 99.92 & 99.43 \\
10 & 100.00 & 99.50 \\
15 & 100.00 & 99.63 \\
20 & 100.00 & 99.67 \\
\hline
\end{tabular}
\end{table}

The proxy reaches 100\% accuracy by $k=10$, exhibiting the same ceiling behavior reported for LINC. This confirms that our implementation faithfully captures LINC-style majority-voting reasoning.

\subsection{Solver-weighted Aggregation Strategy}
\label{sec:aggregation-ablation}

In the main experiments, we adopt majority voting over pruned paths as the aggregation strategy. To examine whether alternative selection mechanism, we conduct a lightweight post-hoc aggregation study using solver-feedback-weighted voting.

\paragraph{Solver-weighted aggregation.}
After \texttt{both} pruning, each path is assigned a weight based on solver feedback:

\begin{itemize}
    \item Paths passing both existence and uniqueness checks receive weight 1.0.
    \item Paths passing only one check (but producing a non-empty answer set) receive weight 0.5.
    \item Empty-answer paths contribute zero vote mass.
\end{itemize}

For multi-answer paths, the path weight is evenly distributed across candidate labels. Final prediction is obtained by summing vote mass per label and selecting the maximum.

\begin{table}[htbp]
\centering
\small
\caption{Majority vs.\ solver-weighted aggregation on AR-LSAT ($k=20$).}
\label{tab:aggregation}
\resizebox{\columnwidth}{!}{
\begin{tabular}{l|cc}
\hline
Model 
& Majority vote (\%) 
& Solver-weighted vote (\%) \\
\hline
GPT-4o & 78.00 & 81.74 \\
GPT-5.2 & 93.91 & 94.35 \\
Gemini-3-flash & 95.65 & 95.65 \\
\hline
\end{tabular}
}
\end{table}

The results are shown in Table~\ref{tab:aggregation}. Solver-weighted aggregation yields additional gains on some backbones (notably +3.74\% on GPT-4o), while remaining comparable on near-ceiling models. This indicates that solver feedback provides useful confidence signals beyond plain majority voting. Anyway, D\&P is modular with respect to aggregation.

\subsection{Ablation study on $T_{DRAFT}$}
\label{sec:draft-temp}
We ran a drafting-temperature ablation on AR-LSAT with GPT-4o. Accuracy peaks at $T_{\text{DRAFT}} = 1.0$ and degrades as temperature increases further, indicating that moderate stochasticity improves diversity while excessive randomness harms faithfulness and stability. These results empirically justify our default choice of $T_{\text{DRAFT}} = 1.0$.
\begin{table}[htbp]
\centering
\caption{Drafting temperature ablation on AR-LSAT with GPT-4o (Acc@k20).}
\resizebox{\columnwidth}{!}{
\begin{tabular}{c|cccccc}
\hline
$T_{\text{DRAFT}}$ & 0.0 & 0.5 & 1.0 & 1.1 & 1.2 & 1.3 \\
\hline
Acc@k20 (\%) & 60.17 & 71.87 & 78.00 & 75.83 & 75.65 & 70.87 \\
\hline
\end{tabular}
}
\end{table}

\subsection{Ablation study on ICL Demonstration Count}
\label{sec:icl-ablation}

We analyze the effect of the number of in-context learning (ICL) demonstrations on AR-LSAT using GPT-4o with $k=20$. We evaluate 0-shot, 1-shot, 3-shot, and 5-shot settings for plan drafting and program generation.

Table~\ref{tab:icl-ablation} reports accuracy and average token usage per path. Token counts are decomposed into average input and output tokens.

\begin{table}[htbp]
\centering
\small
\caption{ICL shot ablation on AR-LSAT (GPT-4o, $k=20$).}
\label{tab:icl-ablation}
\resizebox{\columnwidth}{!}{
\begin{tabular}{c|cccc}
\hline
ICL & Acc@k20 
& \makecell{Avg input\\tokens/path}
& \makecell{Avg output\\tokens/path}
& \makecell{Avg total\\tokens/path} \\
\hline
0 & 55.30 & 1{,}555 & 1{,}574 & 3{,}129 \\
1 & 65.65 & 2{,}684 & 1{,}199 & 3{,}883 \\
3 & 78.00 & 5{,}200 & 1{,}057 & 6{,}257 \\
5 & 78.61 & 8{,}817 & 1{,}078 & 9{,}895 \\
\hline
\end{tabular}
}
\end{table}

Accuracy improves monotonically with additional demonstrations and peaks at 5-shot (78.61\%). However, the gain over 3-shot (78.00\%) is marginal (+0.61 absolute) while total token usage increases substantially.

The additional cost beyond 3-shot is primarily driven by increased input tokens due to longer ICL context, whereas output length remains relatively stable. This indicates diminishing returns in accuracy relative to prompt length expansion. Based on this trade-off, we adopt 3-shot as the default configuration.
\subsection{Ablation on Maximum Repair Depth}
\label{sec:repair-depth}

We study the effect of repair depth in the single-path setting ($k=1$) on AR-LSAT with GPT-4o. Here, $\texttt{max\_repairs} \in \{0,1,2,3\}$ denotes the maximum number of syntax repair rounds allowed after the initial code generation attempt.

\begin{table}[htbp]
\centering
\small
\caption{Effect of maximum repair rounds on AR-LSAT (GPT-4o, $k=1$).}
\label{tab:repair-depth}
\resizebox{\columnwidth}{!}{
\begin{tabular}{c|ccc}
\hline
$\texttt{max\_repairs}$ 
& Acc (\%) 
& ExecRate (\%)
& Acc@Exec (\%)  \\
\hline
0 & 36.09 & 87.83 & 41.09 \\
1 & 40.00 & 96.52 & 41.44 \\
2 & 45.65 & 97.39 & 46.88 \\
3 & 36.52 & 97.83 & 37.33 \\
\hline
\end{tabular}
}
\end{table}

Table~\ref{tab:repair-depth} shows that allowing a small number of repair rounds substantially improves solver executability: \texttt{ExecRate} rises from 87.83\% with no repair to 97.39\% with two repair rounds, and remains high at 97.83\% with three rounds. However, end-to-end accuracy is non-monotonic. Accuracy improves from 36.09\% to 45.65\% when increasing $\texttt{max\_repairs}$ from 0 to 2, but drops to 36.52\% at $\texttt{max\_repairs}=3$. A similar pattern is reflected in \texttt{Acc@Exec}, which peaks at 46.88\% for $\texttt{max\_repairs}=2$ and then declines to 37.33\%.

These results suggest that additional repair rounds can improve executability without necessarily improving the semantic correctness of the resulting formalization. We therefore use $\texttt{max\_repairs}=2$ as the default setting.

\subsection{Failure Analysis}
\label{sec:failure-analysis}
Although pruning substantially improves performance, a portion of AR-LSAT instances remains incorrect under $k=20$ sampling. To better understand the source of these residual errors, we conduct a structured failure analysis under the setting $k=20$ with pruning.

Recall that \texttt{Coverage@k} measures whether at least one correct path appears among $k$, providing a quantitative signal of drafting-stage coverage. To further diagnose failure modes, we categorize the incorrect instances into two top-level groups:

\begin{itemize}
    \item \textbf{Coverage failure:}
    \begin{itemize}
        \item \emph{(A) Empty candidate set after pruning:} All sampled paths are filtered by pruning.
        \item \emph{(B) Non-empty candidate set, but no correct candidate:} At least one path survives pruning, but none is correct.
    \end{itemize}

    \item \textbf{Aggregation failure:} At least one correct candidate survives pruning, but majority voting selects an incorrect answer.
\end{itemize}

Table~\ref{tab:failure-taxonomy-gpt-4} and Table~\ref{tab:failure-taxonomy-gpt-4o} summarizes the distribution of failure categories for GPT-4 and GPT-4o.

\begin{table}[htbp]
\centering
\caption{Failure taxonomy on AR-LSAT ($k=20$, with pruning, GPT-4).}
\label{tab:failure-taxonomy-gpt-4}
\resizebox{\columnwidth}{!}{
\begin{tabular}{lccc}
\hline
Category & Count & \% Fail & \% All \\
\hline
Coverage failure A & 19 & 37.25 & 8.26 \\
Coverage failure B & 16 & 31.37 & 6.96 \\
Aggregation failure & 16 & 31.37 & 6.96 \\
\hline
\end{tabular}
}
\end{table}

\begin{table}[htbp]
\centering
\caption{Failure taxonomy on AR-LSAT ($k=20$, with pruning, GPT-4o).}
\label{tab:failure-taxonomy-gpt-4o}
\resizebox{\columnwidth}{!}{
\begin{tabular}{lccc}
\hline
Category & Count & \% Fail & \% All \\
\hline
Coverage failure A & 24 & 48.00 & 10.43 \\
Coverage failure B & 13 & 26.00 & 5.65 \\
Aggregation failure & 13 & 26.00 & 5.65 \\
\hline
\end{tabular}
}
\end{table}

Across both models, the majority of residual errors arise from  coverage limitations. Aggregation failures constitute a smaller portion of errors. These results suggest that further gains are likely to come from improving drafting diversity and coverage.
\subsection{Corrected AR-LSAT Test Split}
\label{sec:appx-ar-lsat-fixed}

We found annotation issues in the public test set of AR-LSAT and manually audited the incorrect labels.

Starting from the original 230-sample test file, we remove one mismatched sample, \texttt{201510\_3-G\_3\_19}, whose question does not match the context. We also relabel the following eight samples:
\begin{itemize}
    \item \texttt{201310\_3-G\_3\_14}
    \item \texttt{201612\_3-G\_1\_3}
    \item \texttt{201612\_3-G\_1\_4}
    \item \texttt{201612\_3-G\_1\_5}
    \item \texttt{201612\_3-G\_2\_6}
    \item \texttt{201612\_3-G\_2\_8}
    \item \texttt{201612\_3-G\_3\_12}
    \item \texttt{201612\_3-G\_3\_17}
\end{itemize}
The resulting corrected benchmark contains 229 samples. For anonymous review, we provide an anonymous Hugging Face dataset link:
\begin{center}
\url{https://huggingface.co/datasets/anonymous-ar-lsat/ar-lsat-fixed-229}
\end{center}


\subsection{Cost-Normalized Comparison Across Sampling Budgets}
\label{sec:cost-normalized}

To analyze whether the performance gains of D\&P are attributable merely to larger sampling budgets, we compare for CoT-SC~\cite{wang2022self} and D\&P accuracy as a function of $k$ on AR-LSAT with GPT-4o.

For each method, the total token budget per instance is computed as
\[
\text{Tokens per sample} = k \times \text{Avg tokens per path}.
\]

Table~\ref{tab:cost-normalized} reports accuracy and total tokens per sample for $k \in \{1,5,10,15,20\}$.

\begin{table}[htbp]
\centering
\caption{Cost-normalized comparison on AR-LSAT (GPT-4o).}
\label{tab:cost-normalized}
\begin{tabular}{c|cc|cc}
\hline
& \multicolumn{2}{c|}{CoT-SC} & \multicolumn{2}{c}{D\&P} \\
$k$ & Acc@k & Tokens & Acc@k & Tokens \\
\hline
1  & 34.09 & 2{,}008   & 39.87 & 6{,}257   \\
5  & 36.78 & 10{,}040  & 64.04 & 31{,}283  \\
10 & 37.70 & 20{,}081  & 72.17 & 62{,}567  \\
15 & 38.52 & 30{,}121  & 75.91 & 93{,}850  \\
20 & 38.35 & 40{,}162  & 78.00 & 125{,}134 \\
\hline
\end{tabular}
\end{table}

Across all sampling budgets, D\&P consumes approximately $3\times$ more tokens per instance than CoT-SC. However, the accuracy gap persists across the entire $k$ range. Notably, even at $k=20$, where CoT-SC uses over 40k tokens per sample, its accuracy remains below 39\%, whereas D\&P reaches 78\%. This suggests that the improvements are not explained solely by larger sampling budgets.
\subsection{Solver Timeout Configuration and Analysis}
\label{sec:timeout}

We clarify the timeout configuration used for solver execution and analyze its impact. Timeouts are backend-dependent:

\begin{itemize}
    \item Z3Py (AR-LSAT): 30s subprocess timeout per program.
    \item PythonConstraint/CSP (LogicalDeduction): 20s subprocess timeout per program.
    \item PyKe (ProofWriter, PrOntoQA): in-process execution with no external subprocess timeout configured.
\end{itemize}

\paragraph{Observed timeout frequency.}
We measure path-level timeout events under $k=20$. The denominator corresponds to the total number of generated paths (i.e., \#instances $\times k$).

\begin{table}[htbp]
\centering
\small
\caption{Path-level timeout statistics under $k=20$.}
\label{tab:timeout}
\begin{tabular}{lccc}
\hline
Benchmark & Solver & GPT-4o & GPT-4 \\
\hline
AR-LSAT & Z3Py & 1 / 4600 & 2 / 4600 \\
ProofWriter & PyKe & 0 / 12000 & 0 / 12000 \\
PrOntoQA & PyKe & 0 / 10000 & 0 / 10000 \\
LogicalDeduction & CSP & 0 / 6000 & 4 / 6000 \\
\hline
\end{tabular}
\end{table}

Timeout events are extremely rare across all benchmarks. Given the near-zero frequency relative to the total number of evaluated paths, their aggregate impact on ExecRate, Acc@Exec, and final accuracy is negligible.
\subsection{Diversity Measurement via Normalized Answer Entropy}
\label{sec:plan-entropy}

Beyond final accuracy, we report \texttt{ExecRate}, \texttt{Acc@Exec}, and \texttt{Coverage@k} in the main paper, which quantify executability and drafting-stage coverage. To characterize the dispersion of draft answers, we introduce an answer-level entropy statistic computed over valid single-label parsed outputs.

\paragraph{Definition of $H_{\mathrm{avg}}^{\mathrm{norm}}$.}
For each instance $i$, we collect the parsed answers from $k$ draft paths and retain only valid single-label outputs (i.e., exactly one option in $\mathcal{C}$). Let $n_{i,c}$ denote the number of retained paths whose parsed answer equals option $c$. Define
\[
p_{i,c} = \frac{n_{i,c}}{\sum_{c' \in \mathcal{C}} n_{i,c'}}.
\]
The instance-level entropy is
\[
H_i = - \sum_{c \in \mathcal{C}} p_{i,c} \log p_{i,c},
\]
with normalized entropy
\[
H_i^{\mathrm{norm}} = \frac{H_i}{\log |\mathcal{C}|}.
\]
We report the dataset average
\[
H_{\mathrm{avg}}^{\mathrm{norm}} = \frac{1}{N} \sum_{i=1}^{N} H_i^{\mathrm{norm}}.
\]
Under our current implementation on AR-LSAT, pruning enforces the same single-label constraint (existence/uniqueness), and therefore $H_{\mathrm{avg}}^{\mathrm{norm}}$ is identical whether computed before or after pruning.

\paragraph{Results across drafting temperature.}
Table~\ref{tab:entropy-temp} reports \texttt{Acc@k20}, \texttt{PathAcc} (the correct paths over all paths), \texttt{HitRate@k20}, and $H_{\mathrm{avg}}^{\mathrm{norm}}$ for different drafting temperatures $T_{\text{DRAFT}}$.

\begin{table}[htbp]
\centering
\small
\caption{Answer distribution statistics vs.\ drafting temperature on AR-LSAT (GPT-4o, $k=20$; entropy computed on valid single-label paths).}
\label{tab:entropy-temp}
\resizebox{\columnwidth}{!}{
\begin{tabular}{c|cccc}
\hline
$T_{\text{DRAFT}}$ 
& Acc@k20 (\%) 
& PathAcc (\%) 
& HitRate@k20 (\%) 
& $H_{\mathrm{avg}}^{\mathrm{norm}}$ \\
\hline
0.0 & 60.17 & 39.43 & 61.30 & 0.026 \\
0.5 & 71.87 & 39.17 & 75.65 & 0.049 \\
1.0 & 78.00 & 39.13 & 83.91 & 0.068 \\
1.1 & 75.83 & 38.30 & 82.17 & 0.076 \\
1.2 & 75.65 & 36.96 & 80.43 & 0.096 \\
1.3 & 70.87 & 34.10 & 76.42 & 0.102 \\
\hline
\end{tabular}
}
\end{table}

As $T_{\text{DRAFT}}$ increases, $H_{\mathrm{avg}}^{\mathrm{norm}}$ increases monotonically, indicating greater dispersion of answers. Meanwhile, \texttt{PathAcc} decreases, showing that individual paths become less reliable at higher temperatures. In contrast, \texttt{HitRate@k20} first increases and peaks at $T_{\text{DRAFT}}=1.0$, closely aligning with the peak of \texttt{Acc@k20}. This pattern suggests that moderate stochasticity improves coverage. When temperature becomes too high, big dispersion reduces \texttt{HitRate@k20} and consequently lowers \texttt{Acc@k20}.

\subsection{Normalized Prompt Schema}
\label{sec:prompt-schema}
Before presenting the full executable prompts in Appendix~A.3, we describe a normalized prompt schema shared across benchmarks. This abstraction clarifies that performance depends on a fixed structural interface rather than benchmark-specific phrasing.

\paragraph{Unified slot interface.}
All prompts are instantiated from the same slot schema:

\begin{description}
    \item[\texttt{ROLE}] solver-aware assistant specification.
    \item[\texttt{TASK}] reasoning objective (plan, code-from-plan, or syntax repair).
    \item[\texttt{OUTPUT\_FORMAT}] strict output format (answer-only, plan-only or code-only).
    \item[\texttt{ICL\_BLOCK}] fixed 3-shot demonstrations.
    \item[\texttt{INSTANCE}] \{\texttt{context}\}, \{\texttt{question}\}, \{\texttt{answers}\}.
    \item[\texttt{DRAFT}] \{\texttt{plan}\}.
    \item[\texttt{REPAIR\_INPUT}] \{\texttt{code}\}, \{\texttt{syntax\_error}\}.
\end{description}

\paragraph{Prompt families.}
Each method corresponds to a deterministic composition of these slots:

\newcommand{\slotplus}{\ \allowbreak+\ \allowbreak}

\begin{itemize}
  \item {\raggedright \textbf{CoT:}\\
  \texttt{ROLE\slotplus TASK\slotplus OUTPUT\_FORMAT(answer-only)\slotplus ICL\_BLOCK\slotplus INSTANCE}\par}

  \item {\raggedright \textbf{D\&P Draft Generation:}\\
  \texttt{ROLE\slotplus TASK(plan)\slotplus OUTPUT\_FORMAT(plan-only)\slotplus ICL\_BLOCK\slotplus INSTANCE}\par}

  \item {\raggedright \textbf{D\&P Code Generation:}\\
  \texttt{ROLE\slotplus TASK(code-from-plan)\slotplus OUTPUT\_FORMAT(code-only)\slotplus ICL\_BLOCK\slotplus INSTANCE\slotplus DRAFT}\par}

  \item {\raggedright \textbf{D\&P Syntax Repair:}\\
  \texttt{ROLE\slotplus TASK(repair)\slotplus OUTPUT\_FORMAT(code-only)\slotplus REPAIR\_INPUT}\par}
\end{itemize}

\paragraph{Backend adapter.}
Benchmark differences are isolated to the solver layer:
AR-LSAT uses Python + Z3; LogicalDeduction uses Python + \texttt{python-constraint}; PrOntoQA and ProofWriter use Python + PyKe. All other structural slots are shared.
}

\end{document}